# Augmented Reality needle ablation guidance tool for Irreversible Electroporation in the pancreas

Timur Kuzhagaliyev[a,b,c], Neil T. Clancy*[a,b,c], Mirek Janatka[a,b,c], Kevin Tchaka[a,b,c], Francisco Vasconcelos[a,b,d], Matthew J. Clarkson[a,b,d], Kurinchi Gurusamy[e], David J. Hawkes[a,b,d], Brian Davidson[e], Danail Stoyanov[a,b,c]

[a]Wellcome/EPSRC Centre for Interventional & Surgical Sciences (WEISS), University College London, UK; [b]Centre for Medical Image Computing (CMIC), University College London, UK; [c]Department of Computer Science, University College London, UK; [d]Department of Medical Physics and Biomedical Engineering, University College London, UK; [e]Division of Surgery and Interventional Science, UCL Medical School, Royal Free Hospital, University College London, UK

## ABSTRACT

Irreversible electroporation (IRE) is a soft tissue ablation technique suitable for treatment of inoperable tumours in the pancreas. The process involves applying a high voltage electric field to the tissue containing the mass using needle electrodes, leaving cancerous cells irreversibly damaged and vulnerable to apoptosis. Efficacy of the treatment depends heavily on the accuracy of needle placement and requires a high degree of skill from the operator. In this paper, we describe an Augmented Reality (AR) system designed to overcome the challenges associated with planning and guiding the needle insertion process. Our solution, based on the HoloLens (Microsoft, USA) platform, tracks the position of the headset, needle electrodes and ultrasound (US) probe in space. The proof of concept implementation of the system uses this tracking data to render real-time holographic guides on the HoloLens, giving the user insight into the current progress of needle insertion and an indication of the target needle trajectory. The operator's field of view is augmented using visual guides and real-time US feed rendered on a holographic plane, eliminating the need to consult external monitors. Based on these early prototypes, we are aiming to develop a system that will lower the skill level required for IRE while increasing overall accuracy of needle insertion and, hence, the likelihood of successful treatment.

## 1. INTRODUCTION

Irreversible electroporation is a promising therapy for the treatment of inoperable tumours in the pancreas. It involves the placement of two or more needle electrodes into the tissue surrounding the mass followed by the application of a pulsed, high voltage, signal over a period of minutes. The resultant electric field causes nanometer-sized holes to develop in cell membranes within the treatment volume, inducing programmed cell death (apoptosis)[1]. Critical surrounding structures, such as blood vessels and nerve tissue, have been shown to be unaffected by the treatment[2], making it a particularly attractive prospect for otherwise inoperable pancreatic head tumours close to the superior mesenteric artery and vein.

Efficacy of the treatment, however, is dependent on accurate placement of the needles to ensure that the desired target volume is captured. Current approaches involve insertion percutaneously or during open surgery under guidance from pre-operative CT or intraoperative ultrasound imaging[3, 4]. However, this process involves a high degree of skill from the operator to place the electrode at the correct position and angle within the organ, and involves the conceptual challenge of matching images that are two-dimensional representations of 3D volumes (CT, US), with the live surgical field.

The field of image guidance aims to address this challenge, and a considerable body of work has demonstrated its potential for combining and displaying differing modalities in a way that is intuitive and useful for the clinician[5]. This has resulted in several mixed/augmented reality (AR) systems capable of merging standard colour video and medical imaging modalities (CT, MRI, US, gamma) for applications in cardiac surgery[6], needle-biopsy[7] and brain tumour resection[8].

*n.clancy@ucl.ac.uk; surgicalvision.cs.ucl.ac.uk/

We propose an AR system, based on the HoloLens (Microsoft, USA) platform, that will allow the user to scan an organ with a US probe to identify the tumour mass, plan the needle insertion trajectory and target volume, and provide a visual guide during insertion of the needle itself. In this work the IRE needle used in the design of the experiment is the NanoKnife system (AngioDynamics, USA). HoloLens will enable visualization of US images of the tumour, as well as the planned and actual NanoKnife trajectories, on the live surgical field-of-view. An external tracking system, using optical markers, ensures that the coordinate systems of each component can be registered.

In this paper we describe the results of a proof-of-concept study for this system, using a NanoKnife, US probe and a tissue surrogate. A visualization system for merging the planned and live trajectories is demonstrated, and potential future developments are discussed.

## 2. METHODS

### 2.1 Interactive AR guides

HoloLens is equipped with an inertial measurement unit, depth-sensing camera, RGB camera and four infra-red (IR) cameras used for mapping its surroundings, allowing it to determine its position and orientation in space to a high degree of precision[9]. Internally, HoloLens maintains a coordinate system registered with the physical environment around it. This allowed us to register interactive holograms with certain static objects near HoloLens, e.g., registering a hologram representing a needle trajectory with a static target (in future applications, a patient's body). HoloLens is also capable of processing voice commands from the user with input from four microphones on the headset.

We developed our HoloLens application as a Universal Windows Platform (UWP) application using the Unity3d engine. One of the primary functions of the application is to wirelessly communicate with the tracking server and register holograms with their physical counterparts, e.g., superimposing the virtual needle on top of the physical needle to give the user an indication that the object is currently being tracked. The application also allows the user to define a 'planned' trajectory for the needle and subsequently draws visual guides to indicate the displacement and angle offset between this and the current, 'actual', trajectory. The user can interact with the application and adjust settings using either voice commands or hand gestures, eliminating any physical contact and reducing the risk of contamination.

### 2.2 Infra-red optical tracking

In our system, we use a V120:Trio (OptiTrack, USA) equipped with three stereoscopic IR cameras to track optical IR markers attached to the IRE needles and the ultrasound probe (Figure 1), as well as the HoloLens headset itself (Figure 2). The V120:Trio, in turn, interfaces with the Motive motion capture software platform (OptiTrack, USA), which interprets the positions of the optical IR markers in space and allows them to be grouped into rigid bodies.

Broadcasting tracking data from Motive introduces several challenges. First, Motive uses the NatNet SDK to broadcast tracking data, which currently does not provide any libraries for the UWP platform. Secondly, Motive's local coordinate system is right-handed, while Unity3d uses a left-handed coordinate system. To remedy this we developed a TCP server that acts as middleware between Motive and the UWP application on the HoloLens. It extracts information from NatNet frames containing the pose of tracked rigid bodies, performs necessary transformations on the coordinates of rigid bodies and exposes data to HoloLens over Wi-Fi using a simple TCP protocol.

Motive supports grouping multiple optical markers into rigid bodies for tracking purposes. We created a separate rigid body for the HoloLens headset, US probe and an IRE needle. Our UWP application contains 3d models that act as counterparts of these rigid bodies - we run several calibration algorithms on HoloLens so pivots of Motive rigid bodies and Unity3d objects need not match up perfectly.

We use hand-eye calibration to work out the transformation between Motive's and Unity's coordinate systems and hence determine the position and orientation of the OptiTrack cameras in Unity's reference frame. For the IRE needle, we use sphere fitting and circle fitting to calibrate the tip and main axis respectively. For the US feed, we produce two clouds of points (one from the position of the IRE needle in Unity's reference frame, another coordinates of the needle in US feed) and match them up using an absolute orientation algorithm, adjusting the position of the US plane in Unity to match the physical location of the scanned object. This brief calibration sequence only needs to be repeated once during the initial setup of the system. Additional needle calibration with respect to the US plane can also be performed but was omitted for this study and is something we will consider incorporating in the future[10, 11].

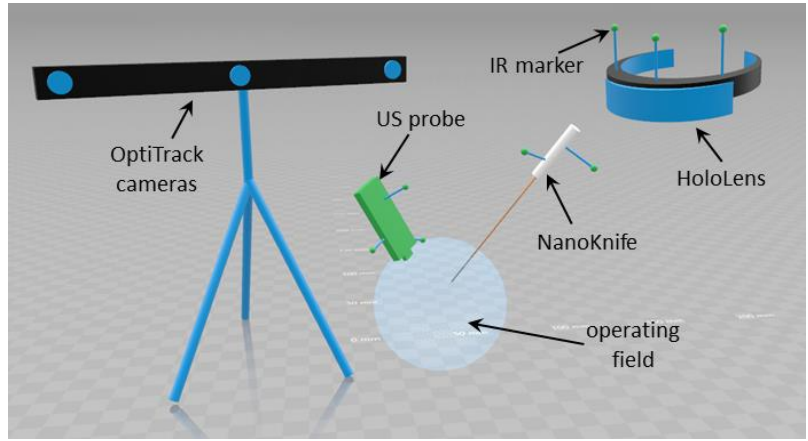

Figure 1. Schematic of the experimental set-up showing the relative positions of the HoloLens headset, surgical instruments and OptiTrack system components.

Once the operator launches the HoloLens application and inputs the address of the TCP server, HoloLens begins receiving tracking data for all registered rigid bodies. This data is used by the UWP application to perform initial pivot calibration of all objects. To compensate for any lag during this phase we maintain a history of the headset's position in space and retrieve the relevant position based on the timestamp of the tracking data. Once this initial calibration is complete, our UWP application fixes the position of the OptiTrack cameras in its coordinate system. This means that the user wearing HoloLens can now move out of cameras' field of view without disrupting the system. Other objects (IRE needle, US probe) must still remain in the cameras' field of view for accurate tracking to be possible. In contrast with using on-board HoloLens sensors for tracking, this approach does not require the needles to remain in the operator's field of view. The tracking server keeps updating HoloLens with real-time tracking data regardless of which direction the user is facing. This, combined with the wireless nature of our system, allows the operator to move around freely without any danger of disrupting the accuracy of tracking.

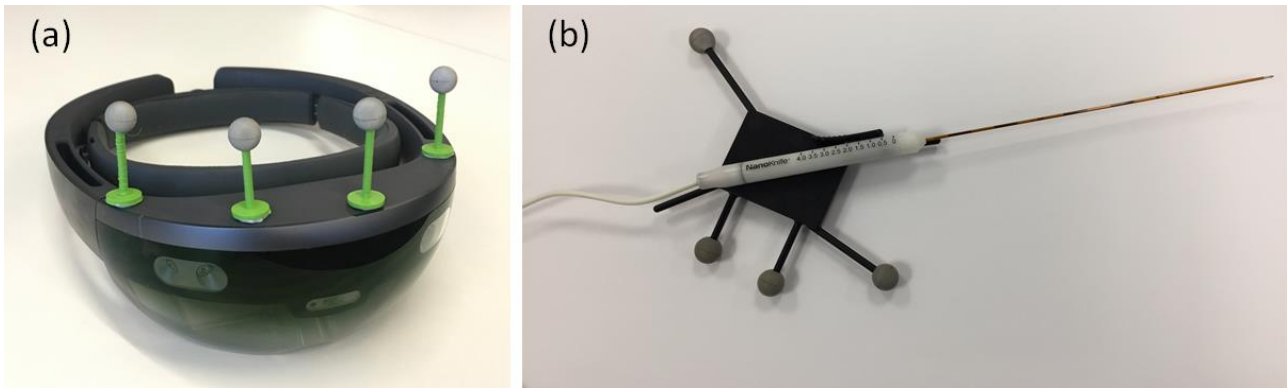

Figure 2. Placement of optical IR markers. (a) HoloLens headset. (b) NanoKnife needle. The slider on the handle controls the active length of the electrode that is exposed to the tissue.

## 3. RESULTS

Figure 3 shows an experiment conducted to render a US feed on a plane registered with the US probe. The application we developed supports streaming of multiple video feeds simultaneously from any generic video source. A video feed can be rendered on a static plane, acting as a holographic monitor, or registered with any tracked object. For this experiment the output of the US system (Ultrasonix; Analogic, USA) was connected to a video capture unit (Pearl; Epiphan Systems, Inc, USA) and then streamed wirelessly to the HoloLens as described in Section 2.

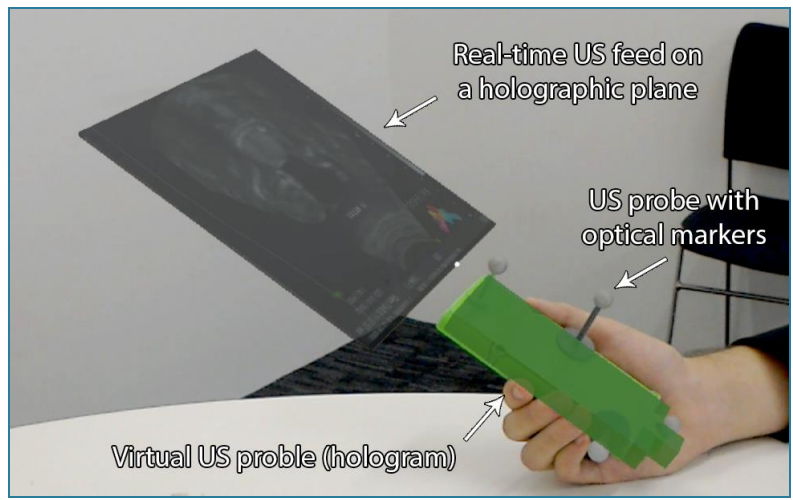

Figure 3. US feed rendered on a plane registered with a US probe. Image taken using HoloLens Mixed Reality Capture application mode. The images would be used to guide IRE probe insertion by visualising the US plane in the direct line-of-sight of the surgeon rather than looking at a remote screen.

Our application allows the operator to input the target needle trajectories, indicating the point of entry into the body and the target tumour. Based on this data, the system draws holographic guides indicating current offset of the needle from the target trajectory. The operator can see an indication of their progress along the target trajectory and magnified displacement from it, as well as current angle offset (Figure 4).

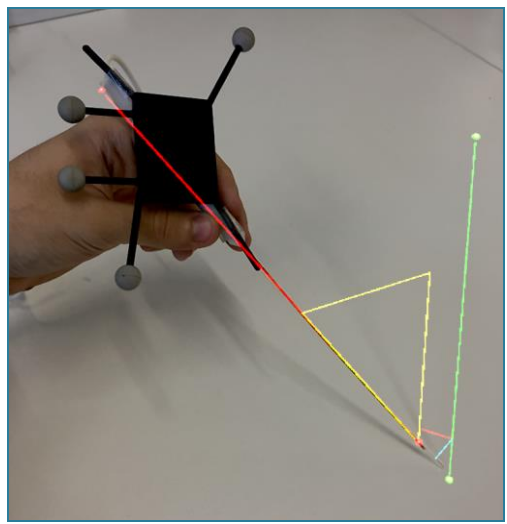

Figure 4. Visual guide for needle placement. Actual trajectory in red and planned trajectory in green with the yellow triangle indicating the deviation between them.

A demonstration of the complete system on an abdominal surgical simulator, with silicone organs (IOUSFAN; Kyoto Kagaku, Japan), is shown in Figure 5. The ultrasound video was broadcast and overlaid as a hologram, in real time, on a plane adjacent to the transducer. A pre-planned needle insertion point in one of the organs is indicated by a green line in the field-of-view, while the NanoKnife needle's current position and deviation from the planned path is also indicated graphically.

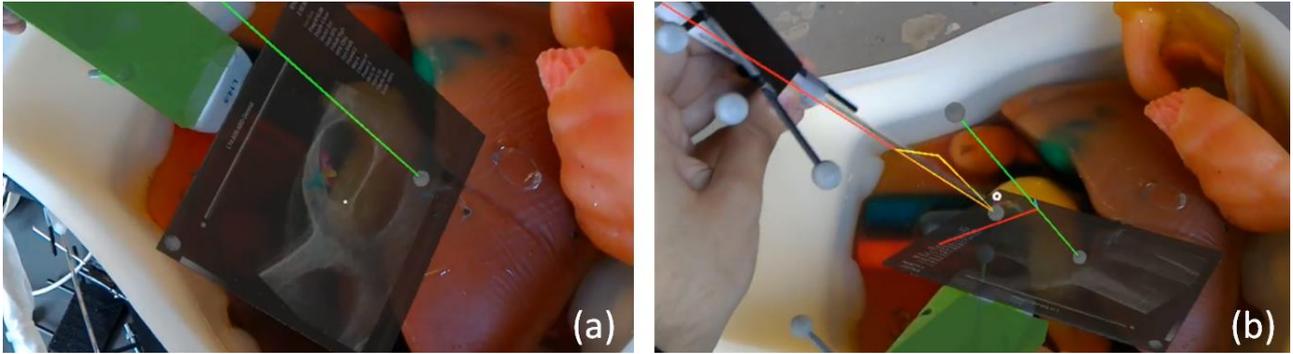

Figure 5. Demonstration of the complete system in an organ phantom. (a) Overlay of the US feed in the plane of the scanner along with the planned needle trajectory in green. (b) Live tracking of the NanoKnife needle and visual guide showing error with respect to the planned trajectory.

## 4. CONCLUSIONS AND FUTURE WORK

The potential to use AR technologies for IRE needle guidance has been demonstrated in this proof-of-concept investigation. We have successfully achieved a qualitatively accurate synchronization of the coordinate systems of tracking software and HoloLens, made possible by a brief initial calibration sequence. The visual guides superimposed on top of the needles provide insight on the current progress of the electrode insertion, and streaming a US feed onto a holographic plane adds more useful information to the live surgical field-of-view. Our system is likely to ease the technical challenges associated with needle placement by providing an intuitive way to visualize the trajectories. Making all of the necessary data available in operator's field of view has the potential to significantly decrease the time required to perform needle electrode insertion, reduce the risk of needle misplacement and increase the effectiveness of IRE therapy through accurate electrode placement. Future experiments will be conducted to quantify this accuracy.

Future developments might involve combining the current needle tracking mechanism with existing US needle guidance solutions, such as electromagnetic tracking approaches[12]. The live US feed could be analysed to extract precise positions of the needles inside the body of the patient. Since IR tracking provides precise position of the US probe relative to the HoloLens headset, the data extracted from US can be used to further increase the accuracy of visual guides. Another additional feature would be to sync and calibrate data together from the US and the HoloLens[13].

At the time of writing HoloLens does not provide any reliable application programming interfaces (APIs) for tracking moving objects (except basic hand gestures) and it is also not possible to access raw IR sensor data from its four internal IR cameras for further processing. To avoid using external tools, one could implement a tracking routine using the feed from the on-board RGB camera and object recognition algorithms. In an attempt to achieve that, our early experiments involved using the Vuforia (PTC, Inc., USA) AR software development kit to process the RGB feed and recognize image patterns attached to the US probe. However, the limited field-of-view of the RGB camera, tracking lag due to poor performance on HoloLens and inability to track small patterns rendered the system unfit for practical use. As a result, the optical IR tracking approach was chosen instead.

This limitation might disappear once Microsoft releases HoloLens computer vision APIs, allowing direct access to raw sensor data. This would open up the possibility of performing all of the necessary tracking using HoloLens alone. As discussed earlier, this approach has some disadvantages but it will significantly decrease the overall size of the system.

## ACKNOWLEDGEMENTS

The work was funded by a Wellcome Trust Pathfinder award (201080/Z/16/Z) and by the EPSRC (EP/N013220/1, EP/N022750/1, EP/N027078/1, NS/A000027/1). The authors would like to thank AngioDynamics for lending the NanoKnife system for this project.